\documentclass[10pt,twocolumn,letterpaper]{article}

\usepackage{iccv}              

\definecolor{iccvblue}{rgb}{0.21,0.49,0.74}
\usepackage[pagebackref,breaklinks,colorlinks,allcolors=iccvblue]{hyperref}

\usepackage{multirow}

\def\paperID{*****} 
\def\confName{ICCV}
\def\confYear{2025}

\title{Gather-Scatter Mamba: \\Accelerating Propagation with Efficient State Space Model}

\author{
Hyun-kyu Ko$^{1}$ \quad
Youbin Kim$^{2}$ \quad
Jihyeon Park$^{1}$ \quad
Dongheok Park$^{2}$ \quad
Gyeongjin Kang$^{1}$ \quad \\
Wonjun Cho$^{3}$ \quad
Hyung Yi$^{3}$ \quad
Eunbyung Park$^{4}$\thanks{Corresponding author} \\
$^{1}$Department of Electrical and Computer Engineering, Sungkyunkwan University \\
$^{2}$Department of Artificial Intelligence, Sungkyunkwan University \\
$^{3}$Hanwha Systems, Republic of Korea \\
$^{4}$Department of Artificial Intelligence, Yonsei University \\
{\tt\small \{laniko, ybin108, fairytale, leao8869, ggggjin99\}@skku.edu} \\
{\tt\small \{wonjun78.cho, hyung.yi\}@hanwha.com \quad epark@yonsei.ac.kr}
}

\begin{document}
\maketitle

\begin{abstract}
State Space Models (SSMs)—most notably RNNs—have historically played a central role in sequential modeling. Although attention mechanisms such as Transformers have since dominated due to their ability to model global context, their quadratic complexity and limited scalability make them less suited for long sequences. Video super-resolution (VSR) methods have traditionally relied on recurrent architectures to propagate features across frames. However, such approaches suffer from well-known issues including vanishing gradients, lack of parallelism, and slow inference speed. Recent advances in selective SSMs like Mamba~\citep{gu2023mamba} offer a compelling alternative: by enabling input-dependent state transitions with linear-time complexity, Mamba mitigates these issues while maintaining strong long-range modeling capabilities.
Despite this potential, Mamba alone struggles to capture fine-grained spatial dependencies due to its causal nature and lack of explicit context aggregation. To address this, we propose a hybrid architecture that combines shifted window self-attention for spatial context aggregation with Mamba-based selective scanning for efficient temporal propagation. Furthermore, we introduce Gather-Scatter Mamba (GSM), an alignment-aware mechanism that warps features toward a center anchor frame within the temporal window before Mamba propagation and scatters them back afterward, effectively reducing occlusion artifacts and ensuring effective redistribution of aggregated information across all frames.

The official implementation is provided at: \url{https://github.com/Ko-Lani/GSMamba}.

\end{abstract}

\section{Introduction}\label{introduction}

Long before the rise of Transformer architectures~\citep{vaswani2017attention}, recurrent neural networks (RNNs)\citep{elman1990finding} dominated sequence modeling, particularly in early video understanding\citep{yue2015beyond, srivastava2015unsupervised} and video sequence generation~\citep{donahue2015long, venugopalan2015sequence}. However, the inherent limitations of RNNs—vanishing gradients~\citep{bengio1994learning} and limited parallel computation—ultimately led to their decline. Transformers~\citep{vaswani2017attention, liu2021swin, dao2022flashattention} introduced a revolutionary paradigm based on parallel attention mechanisms, now widely adopted across various domains. Despite this success, Transformers' quadratic computational complexity significantly restricts their practicality in video tasks, where long-range temporal modeling is essential.

Temporal modeling remains especially critical in video super-resolution (VSR), which relies on accurately aggregating details from neighboring frames to restore high-frequency visual content. Due to computational constraints, most recent VSR models~\citep{chan2021basicvsr, chan2022basicvsr++, liang2022recurrent, shi2022rethinking, xu2024enhancing} employ recurrent propagation, maintaining explicit hidden states that propagate forward in time. However, these approaches inherently suffer from two major drawbacks. First, the strictly causal nature of propagation limits the model's ability to leverage future frames, even when bidirectional strategies (forward and backward passes) are employed. Second, supporting frames used for propagation are typically discarded immediately afterward, despite the substantial computational cost of extracting their features.

A recent breakthrough in sequence modeling is Mamba~\citep{gu2023mamba}, a structured state space model (S4)~\citep{gu2021efficiently} that employs dynamic, input-dependent transitions. Unlike Transformers, Mamba has linear computational complexity with respect to sequence length, making it highly suitable for long-sequence tasks like video processing. Furthermore, Mamba supports parallel computation while maintaining strong long-range dependency modeling capabilities. Although promising, applying Mamba directly to VSR is nontrivial. 
Previous works in vision~\citep{liu2024vmamba, guo2024mambair, li2024videomamba, zhang20242dmamba} 
have reported that naïve temporal-first scanning can hurt performance, 
and thus typically adopt spatial-first or hybrid spatial–temporal scanning strategies (e.g., Hilbert curves) instead.

To address these challenges, we propose Gather-Scatter Mamba, a novel VSR framework that leverages alignment-aware temporal propagation and symmetric residual redistribution. At the heart of our method lies a gather-scatter strategy: During the gather phase, neighboring frames are aligned to each anchor frame via warping, and the aligned features are temporally flattened and passed through a State Space Model (Mamba) for efficient long-range temporal modeling. In contrast to previous methods that update only the anchor, our scatter phase performs explicit residual redistribution, warping the output residuals from the anchor frame back to each supporting frame. This allows all frames in the temporal window to be enhanced jointly, enabling joint refinement of all frames in the temporal window and maximizing the use of computed features.

Unlike prior sliding-window approaches that treat past frames merely as support for the current frame, we update all frames within the window jointly by centering the anchor and aligning neighbors toward it. This removes the strict past–present distinction and allows information to flow symmetrically from both directions. The resulting design shortens alignment paths, reduces warping error, and enables balanced feature aggregation across the entire temporal window. Although the window still advances sequentially, we additionally perform a backward pass over the sequence, 
ensuring bidirectional propagation and maximizing information reuse across all frames.

Our contributions can be summarized as follows:

\begin{itemize}
\item We propose Gather-Scatter Mamba, the first VSR framework to integrate Mamba~\citep{gu2023mamba} for temporal propagation, enabling long-range temporal modeling with linear complexity.
\item Unlike prior Mamba usage in video models, we introduce a gather-and-align step that explicitly aligns neighboring frames before temporal-first scanning, allowing Mamba to robustly capture long-range dependencies while overcoming spatial misalignments between frames.
\item We introduce a residual redistribution (scatter) mechanism that updates all frames in a temporal window, maximizing efficiency and improving restoration consistency.
\item We replace conventional forward-anchored propagation with a center-anchored propagation scheme that symmetrically leverages past and future frames, enabling robust alignment and global bidirectional information flow across the entire sequence.
\end{itemize}

\begin{figure*}
    \centering
    \includegraphics[width=\linewidth]{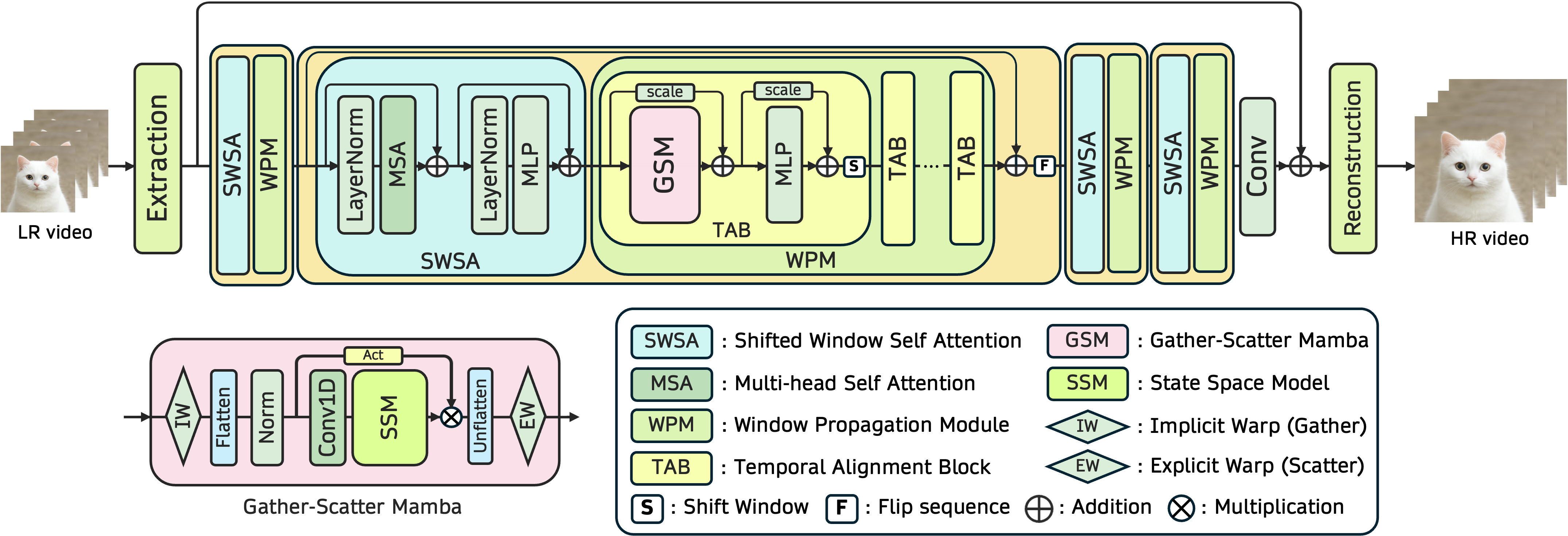}
    \caption{Overall architecture of the proposed Gather-Scatter Mamba (GSM). 
    Given a low-resolution input sequence, local spatial refinement is first performed using shifted window self-attention (SWSA). 
    Temporal propagation is then carried out by the window propagation module (WPM). Within each WPM, the GSM block first \textit{gathers} features by aligning all frames to an anchor frame, processes the aligned features using Mamba's directionally selective scanning, and then \textit{scatters} the updated features back to their original temporal locations.}
\end{figure*}

\section{Related Works}\label{related_works}
\subsection{Video Super-Resolution}
Video Super-Resolution (VSR) aims to recover high-resolution frames from low-resolution video by leveraging both spatial details and temporal redundancy. Unlike single-image SR, VSR must deal with motion, occlusion, and alignment—making it both more powerful and more challenging.
Early models for VSR often adopted recurrent structures to handle the sequential nature of video. Recurrent neural networks (RNNs)~\citep{elman1990finding} provided a natural way to propagate temporal information across frames, enabling temporal consistency and motion-aware enhancement. These models aimed to accumulate context over time, but were fundamentally limited by vanishing gradients and the inefficiency of sequential computation. As a result, they struggled to scale to longer sequences or high-resolution inputs.
With the advent of Transformers~\citep{vaswani2017attention}, VSR models began to shift toward architectures that process multiple frames simultaneously. Models such as VSRT~\citep{cao2021video} and VRT~\citep{liang2024vrt} adopt self-attention to model spatiotemporal dependencies across frames, achieving strong performance through global context aggregation. However, the computational cost of attending to all tokens across multiple high-resolution frames is substantial, making such models difficult to scale in practice.
To manage the high computational cost of full attention across multiple high-resolution frames, many recent models adopt a more practical alternative: propagating features across frames in a recurrent manner using attention or convolutional modules with limited temporal receptive fields. This strategy has become the dominant paradigm in modern VSR, with representative models including BasicVSR~\citep{chan2021basicvsr}, BasicVSR++\citep{chan2022basicvsr++}, RVRT\citep{liang2022recurrent}, and IART~\citep{xu2024enhancing}, all leveraging frame-to-frame propagation to balance efficiency and performance. This paradigm has extended its influence on 3D super-resolution~\citep{shen2024supergaussian, ko2025sequence}, where maintaining multiview consistency across frames is a main challenge.

\subsection{State Space Models}
State Space Models (SSMs)~\citep{hamilton1994state} have long been used in control and signal processing to model temporal dynamics. In deep learning, structured SSMs have emerged as an alternative to RNNs and Transformers, aiming to combine long-range modeling with improved efficiency.
The Structured State Space Sequence model (S4)~\citep{gu2021efficiently} introduced a parameterized kernel derived from linear dynamics, enabling efficient long-range sequence modeling. Follow-up works such as S4D~\citep{gu2022parameterization} and DSS~\citep{mehta2022long} improved stability and generalization, but remained complex and hardware-unfriendly.

Mamba~\citep{gu2023mamba} simplified SSMs through a selective scan mechanism that enables input-dependent transitions, allowing content-aware information routing while maintaining linear complexity. This design supports key capabilities like associative recall and has inspired extensions in vision domains~\citep{liu2024vmamba, huang2024localmamba, yang2024plainmamba, li2024videomamba, gao2024matten, teng2024dim, guo2024mambair, hatamizadeh2024mambavision}, where adaptive spatiotemporal modeling is crucial.

\section{Method}\label{method}
\subsection{Preliminaries}\label{method:preliminaries}
State Space Models (SSMs) are grounded in continuous-time linear time-invariant (LTI) systems, and are traditionally described by the following set of ordinary differential equations (ODEs):

\begin{equation}
\frac{dh(t)}{dt} = Ah(t) + Bx(t), \quad y(t) = Ch(t)
\end{equation}

where $A \in \mathbb{R}^{N \times N}$, $B \in \mathbb{R}^{N \times 1}$, and $C \in \mathbb{R}^{1 \times N}$. This formulation models the evolution of a hidden state $h(t)$ driven by an input signal $x(t)$, producing output $y(t)$.

To apply this system in discrete settings, Zero-Order Hold (ZOH) discretization is commonly used. This leads to the recurrence formulation:

\begin{equation}
h_k = \bar{A} h_{k-1} + \bar{B} x_k, \quad y_k = C h_k
\end{equation}

where the discretized matrices are defined as:

\begin{align}
\bar{A} &= \exp(\Delta A), \\
\bar{B} &= (\Delta A)^{-1} (\exp(\Delta A) - I)\Delta B
\end{align}

Here, $\Delta$ denotes the timestep or step size, which controls how much the model forgets the previous hidden state and incorporates the current input.

The output can also be reformulated in terms of a convolutional kernel:

\begin{equation}
y = x * K, \quad \text{where } K = [C\bar{B},\ C\bar{A}\bar{B},\ \ldots,\ C\bar{A}^k\bar{B}]
\end{equation}

Recent work~\citep{gu2023mamba} introduces Mamba, which challenges the limitations of SSMs with input-invariant parameters $\bar{A}, \bar{B}, C$. Instead, Mamba proposes a selective scan mechanism that allows these parameters to be dynamically generated from the input sequence, enabling fine-grained control over hidden state updates. Specifically, the model uses:
\begin{equation}
B = S_B(x), \quad C = S_C(x), \quad \Delta = \mathrm{softmax}(\theta + S_{\Delta}(x))
\end{equation}
The discrete recurrence parameters are then recomputed as:
\begin{equation}
\bar{A},\ \bar{B} \leftarrow \text{discretize}(\Delta, A, B)
\end{equation}
By allowing the timestep $\Delta$ to be input-dependent, Mamba essentially introduces per-token forget gates, enabling richer and more selective information flow compared to static SSMs.

\begin{figure*}
    \centering
    \includegraphics[width=\linewidth]{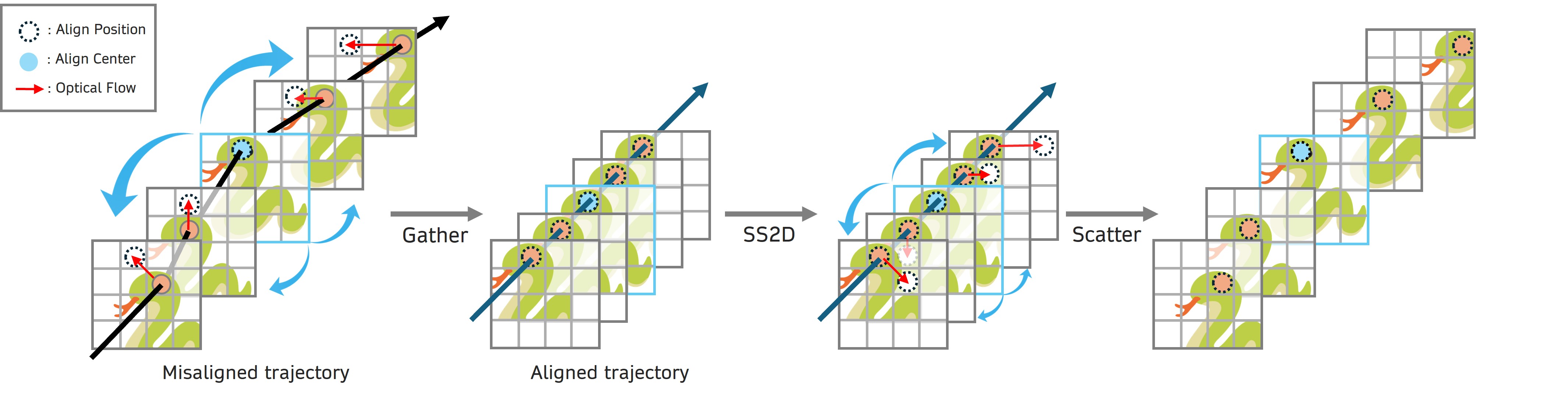}
    \caption{Overview of the proposed Gather-Scatter mechanism.
Misaligned trajectories across frames are first temporally aligned toward the anchor frame (align center) using optical flow (Gather).
The aligned features are then flattened in temporal-first order and processed by SS2D (Mamba) for long-range temporal modeling.
Finally, the output residuals are inversely warped back to their original frame positions (Scatter), updating all frames within the window.}
    \label{fig:gather-scatter}
\end{figure*}

\subsection{Overall Architecture}
Given a low-resolution input video sequence $I_t^{\text{LR}} \in \mathbb{R}^{T \times H \times W \times C}$, the objective is to reconstruct the corresponding high-resolution sequence $I_t^{\text{SR}} \in \mathbb{R}^{T \times sH \times sW \times C}$, where $T$ is the temporal length, $H$ and $W$ are the spatial dimensions, $C$ is the number of channels, and $s$ denotes the upsampling factor.

While image and video super-resolution share similar restoration goals, the key distinction lies in the temporal dimension. Propagating information across time is crucial for video super-resolution, but it often incurs high computational cost. As a result, many existing methods rely on recurrent propagation schemes to balance performance and efficiency.

To address this, we propose a two-stage framework that explicitly decouples spatial refinement and temporal propagation. Our architecture first performs local feature refinement within each frame using shifted window self-attention (SWSA), which enables efficient modeling of intra-frame dependencies with limited receptive fields.

Following spatial refinement, we propagate information temporally using a window-based propagation module (WPM). Within each temporal window, we align all frames to a designated anchor frame through a gather-and-scatter mechanism. In the gather stage, each frame in the window is temporally aligned to the anchor, and the aligned tokens are flattened along the temporal axis. These tokens are then updated via Mamba's directionally selective scanning. In the scatter stage, the updated residuals are warped back to their original temporal locations and aggregated with the local features. This process is repeated by shifting the temporal window across the sequence in both forward and backward directions, enabling bidirectional propagation of temporal information throughout the entire video.

\subsection{Mamba for Video Super-resolution}
Recurrent architectures~\citep{elman1990finding, hochreiter1997long} have been the dominant paradigm in video super-resolution~\citep{tao2017detail, chan2021basicvsr, chan2022basicvsr++, shi2022rethinking, xu2024enhancing}, where each frame is treated as a timestep and features are propagated sequentially. 
However, such approaches are computationally expensive: the recurrent nature enforces strictly sequential propagation, preventing parallelization across frames, and the widespread use of Transformer-based feature extractors incurs quadratic complexity in spatiotemporal dimensions, making both training and inference prohibitively slow for long video sequences.

To address these limitations, we introduce a Mamba-based framework for video super-resolution. Mamba is a structured state space model that enables efficient parallel processing with linear-time complexity, making it well-suited for long-range temporal modeling. To the best of our knowledge, this is the first work to replace temporal propagation with Mamba in the context of video super-resolution.

Whereas existing methods propagate features sequentially across frames, our Mamba-based framework processes the temporal dimension in parallel. Its linear-time design yields a much larger effective temporal receptive field than Transformer-based counterparts, and selective scanning ensures that information from distant frames is preserved and utilized effectively in restoration.

\begin{figure}
    \centering
    \includegraphics[width=\linewidth]{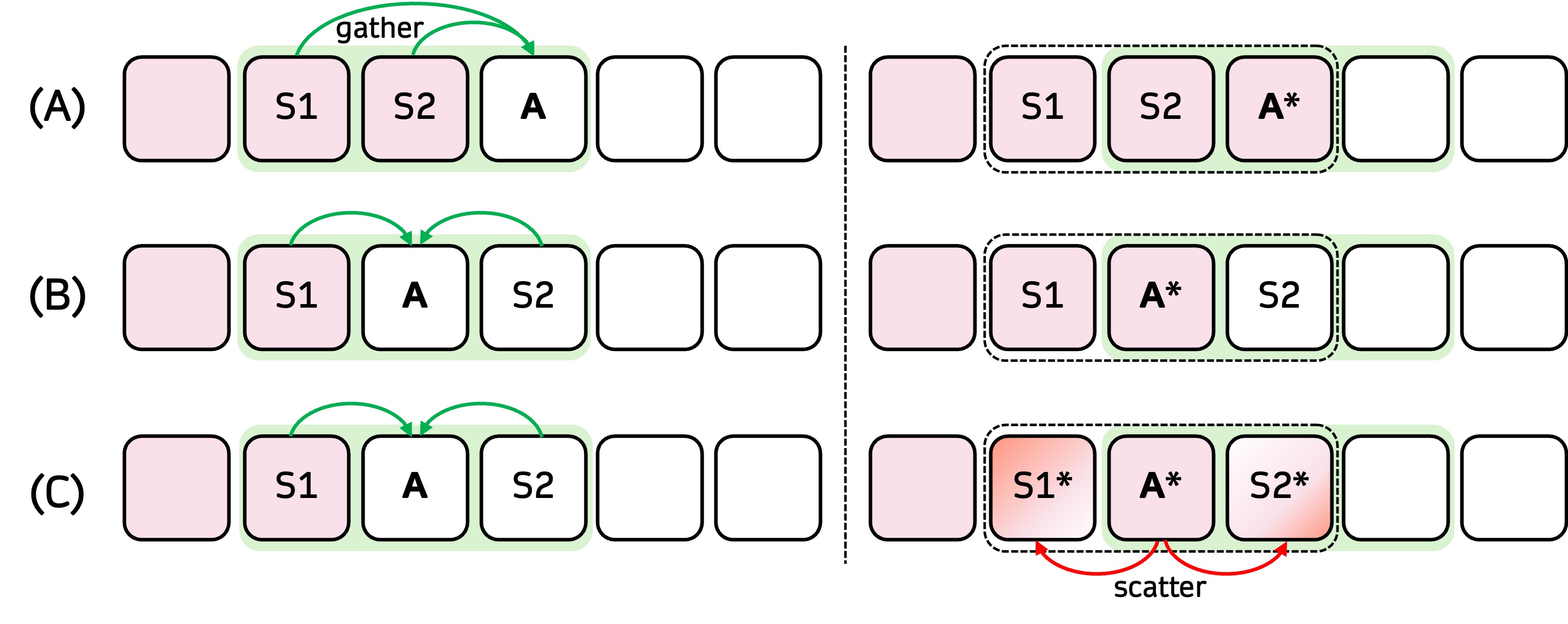}
    \caption{
    Comparison of windowed propagation strategies.
    (A) \textit{Forward-anchored propagation}: supporting frames \textbf{S1}, \textbf{S2} are aligned toward the anchor \textbf{A} located at the end of the window, and only the anchor is updated.
    (B) \textit{Center-anchored propagation}: supporting frames are aligned toward the anchor placed at the center of the window, reducing alignment path length and improving feature aggregation.
    (C) \textit{Center-anchored + Scatter (ours)}: residuals aligned at the center are redistributed back to all supporting frames, enabling joint enhancement of the entire window.
    }
    \label{fig:windowed_propagation}
\end{figure}

\subsection{Gather-Scatter Mamba}
Effective temporal propagation in video restoration critically depends on accurate alignment and efficient reuse of features across frames. Despite the recent successes of Mamba-based architectures~\citep{li2024videomamba, chen2024video, yang2024vivim}, directly applying them to video remains challenging due to their inherent sensitivity to spatial misalignment.

Existing Mamba-based video models predominantly adopt spatial-first scanning~\citep{li2024videomamba, ren2025vamba, wu2024rainmamba} rather than temporal-first scanning, as experiments consistently report inferior performance with the temporal-first approach. This outcome is somewhat counterintuitive, considering that capturing long-range temporal dependencies should theoretically help mitigate local misalignments. A plausible explanation lies in the structural characteristics of Mamba: since it processes video frames as flattened one-dimensional sequences, even minor spatial misalignments between frames translate into substantial positional distortions in the sequence dimension. Consequently, features that are spatially adjacent or semantically similar in 2D may become distant in the 1D sequence, severely hindering the model's ability to associate relevant information. Unlike Transformer-based architectures—which mitigate misalignment issues through local attention windows and flexible token association~\citep{shi2022rethinking}—Mamba strictly relies on sequential adjacency~\citep{huang2024localmamba}.

A conventional approach to temporal feature propagation is forward-anchored propagation (fig.~\ref{fig:windowed_propagation}(A)), which aggregates information from previous frames via optical-flow-based alignment. Formally, given a video frame sequence \(\{x_i\}\) and corresponding features \(f_i^j\) at propagation step \(j\), the forward-anchored residual for frame \(i\) can be computed as:
\begin{equation}
r_i^j = \Phi\!\left(f_i^{j-1},\; W(f_{i-1}^j,\; \mathcal{O}_{i \rightarrow i-1}),\; W(f_{i-2}^j,\; \mathcal{O}_{i \rightarrow i-2})\right),
\end{equation}
where $W(\cdot, \cdot)$ denotes backward warping of a supporting feature $f_k$ toward the reference feature $f_i$ 
using the optical flow $\mathcal{O}_{i \rightarrow k}$ estimated between frame $i$ and frame $k$. Here, \(\Phi(\cdot)\) denotes a feature fusion module, such as a Transformer or Mamba block, that integrates the aligned features and generates a residual update. This strategy, however, neglects useful future context available in subsequent frames.

To leverage bidirectional context, center-anchored propagation incorporates both past and future frames (fig.~\ref{fig:windowed_propagation}(B)):
\begin{equation}
r_i^j = \Phi\!\left(f_i^{j-1},\; W(f_{i-1}^j,\; \mathcal{O}_{i \rightarrow i-1}),\; W(f_{i+1}^{j-1},\; \mathcal{O}_{i \rightarrow i+1})\right).
\end{equation}

Compared to forward-anchored propagation, which relies solely on distant past frames and thus suffers from large temporal gaps and weaker spatial overlap, center-anchored propagation mitigates this issue by symmetrically leveraging nearby past and future frames (Figure~\ref{fig:align_anchor}.) However, it still suffers from another inefficiency: the residuals generated from supporting frames are typically discarded after use, wasting significant computational resources.

To overcome the aforementioned limitations—including mamba's sensitivity to misalignment, suboptimal temporal scanning, and inefficient residual utilization—we propose Gather-Scatter Mamba, composed of two complementary steps:

Prior to Mamba processing, we employ optical-flow-based warping~\citep{ranjan2017optical}
to align features from neighboring frames to the reference frame, which is crucial because Mamba’s 1D sequential scanning is highly sensitive to spatial misalignment (Gather phase):

\begin{equation}
\begin{aligned}
\hat{r}_{k} &= W(\hat{r}_{k \rightarrow i},\; \mathcal{O}_{k \rightarrow i}), \\
f_{k}^{j} &= f_{k}^{j-1} + \hat{r}_{k}, \\
k &\in \left\{ i-\tfrac{K-1}{2},\; \ldots,\; i+\tfrac{K-1}{2} \right\}.
\end{aligned}
\label{eq:scatter}
\end{equation}

The aligned features are then stacked into a 4D tensor and flattened along the temporal dimension
to form a time-major 1D sequence suitable for Mamba processing:

\begin{equation}
\begin{aligned}
\mathbf{G}_i &\in \mathbb{R}^{K \times H \times W \times C}, \\
\tilde{\mathbf{G}}_i &= \text{Flatten}_{\mathrm{temp}}(\mathbf{G}_i) 
   \in \mathbb{R}^{(H \cdot W \cdot K) \times C}, \\
\hat{\mathbf{G}}_i &= \mathcal{M}(\tilde{\mathbf{G}}_i).
\end{aligned}
\label{eq:mamba}
\end{equation}

where $\mathcal{M}$ denotes the Mamba selective scan and ${K}$ denotes the number of frames considered in the local temporal window.

The output sequence is then reshaped back to the spatio-temporal layout and split into per-frame residuals:
\begin{equation}
\{\hat{r}_{k \rightarrow i}\}_{k} = 
\text{Reshape}_{\mathrm{temp}}^{-1}(\hat{\mathbf{G}}_i), 
\quad \hat{r}_{k \rightarrow i} \in \mathbb{R}^{H \times W \times C}.
\label{eq:reshape}
\end{equation}

Finally, the residuals are inversely warped (Scatter phase) and used to update the supporting frames:

\begin{figure}
    \centering
    \includegraphics[width=\linewidth]{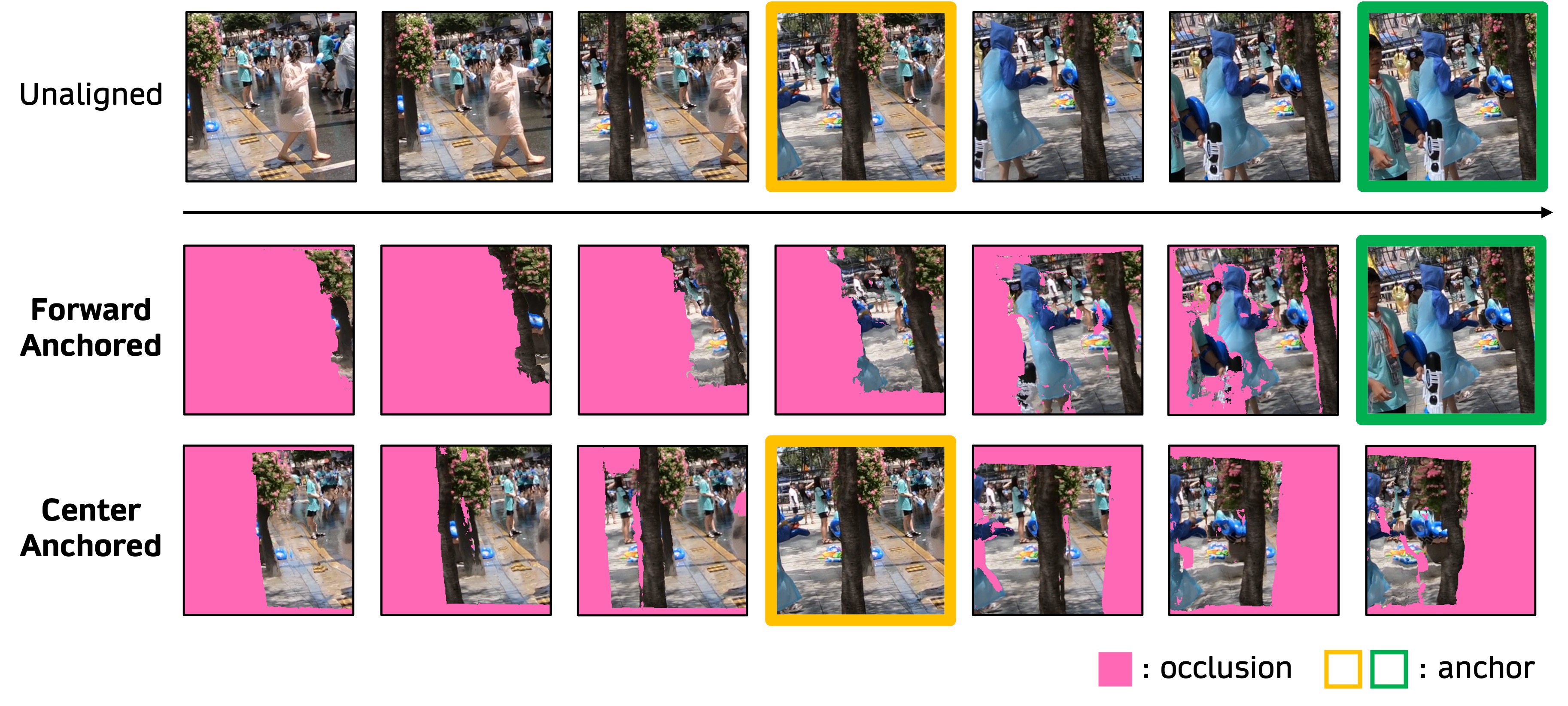}
    \caption{Occlusion comparison between forward-anchored and center-anchored approaches. The center-anchored strategy leverages information from adjacent frames, leading to fewer occluded regions. This reduction in occlusion enables more reliable reconstruction of the anchor frame and improves alignment quality.}
    \label{fig:align_anchor}
\end{figure}

\begin{table*}[t]
\centering
\small
\renewcommand{\arraystretch}{1.2}
\begin{tabular}{l|c|c|c|cc|cc|cc}
\toprule
\multirow{2}{*}{Method} & \multirow{2}{*}{\shortstack{Params\\(M)}} & \multirow{2}{*}{\shortstack{FLOPs\\(T)}} & \multirow{2}{*}{\shortstack{Frames\\REDS/Vimeo}} & \multicolumn{2}{c|}{REDS4} & \multicolumn{2}{c|}{Vimeo-90K-T} & \multicolumn{2}{c}{Vid4} \\
 & & & & PSNR & SSIM & PSNR & SSIM & PSNR & SSIM \\
\midrule
TOFlow & - & - & 5/7 & 27.98 & 0.7990 & 33.08 & 0.9054 & 25.89 & 0.7651 \\
EDVR & 20.6 & 2.95 & 5/7 & 31.09 & 0.8800 & 37.61 & 0.9489 & 27.35 & 0.8264 \\
VSR-T & 32.6 & - & 5/7 & 31.19 & 0.8815 & 37.71 & 0.9494 & 27.36 & 0.8258 \\
BasicVSR & 6.3 & 0.33 & 15/14 & 31.42 & 0.8909 & 37.18 & 0.9450 & 27.24 & 0.8251 \\
IconVSR & 8.7 & 0.51 & 15/14 & 31.67 & 0.8948 & 37.47 & 0.9476 & 27.39 & 0.8279 \\
\midrule
BasicVSR++ & 7.3 & 0.39 & 16/14 & 32.13 & 0.8990 & 37.79 & 0.9500 & 27.79 & 0.8400 \\
VRT & 30.7 & 1.37 & 16/7 & 32.19 & 0.9006 & \underline{38.20} & \underline{0.9530} & 27.93 & 0.8425 \\
RVRT & 10.8 & 2.21 & 16/14 & 32.53 & 0.9078 & 38.15 & 0.9527 & 27.89 & \underline{0.8482} \\
IART & 13.4 & 2.51 & 16/14 & \textbf{32.90} & \textbf{0.9138} & 38.14 & 0.9528 & \textbf{28.26} & \textbf{0.8517} \\
\textbf{GSMamba} & 10.5 & 1.52 & 16/14 & \underline{32.69} & \underline{0.9105} & \textbf{38.25} & \textbf{0.9534} & \underline{28.03} & 0.8462 \\
\bottomrule
\end{tabular}
\caption{Quantitative comparison with the state-of-the-art methods on REDS4 \citep{nah2019ntire}, Vimeo-90K-T \citep{xue2019video}, and Vid4 \citep{liu2013bayesian} datasets for 4$\times$ upsampling. Our GSMamba achieves state-of-the-art performance on Vimeo-90K-T and competitive results on REDS4 and Vid4, while using fewer parameters and FLOPs than existing methods.}
\label{tab:vsr_comparison}
\end{table*}

\begin{equation}
\begin{aligned}
\hat{r}_{k} &= W(\hat{r}_{k \rightarrow i},\; \mathcal{O}_{k \rightarrow i}), \\
f_{k}^{j} &= f_{k}^{j-1} + \hat{r}_{k}, \\
k &\in \left\{ i-\tfrac{K-1}{2},\; \ldots,\; i+\tfrac{K-1}{2} \right\}.
\end{aligned}
\label{eq:scatter}
\end{equation}

This gather–scatter approach ensures spatial correspondences remain intact, significantly alleviating Mamba's inherent sensitivity to misalignment. Moreover, residual reuse (scatter) efficiently recycles valuable intermediate computations, 
enriching the feature representations of supporting frames for subsequent propagation stages. After the scatter phase, the anchor frame index is shifted to $i+1$ and the process is repeated, allowing information to propagate across all frames in a sliding-window manner.

\section{Experiments}\label{experiments}
\subsection{Dataset}
Following prior works~\citep{chan2021basicvsr, chan2022basicvsr++, liang2022recurrent, xu2024enhancing}, we use REDS~\citep{nah2019ntire} and Vimeo-90K~\citep{xue2019video} for training, and evaluate on REDS4, Vimeo-90K-T, and Vid4~\citep{liu2013bayesian}.
REDS provides $\times4$ bicubic downsampled frames with $1280 \times 720$ resolution, which we directly use for training.
Vimeo-90K consists of $448 \times 256$ resolution 7-frame clips; we generate $\times4$ bicubic low-resolution inputs using the MATLAB function from BasicVSR~\citep{chan2021basicvsr} for compatibility.
For Vid4, we evaluate the Vimeo-90K-trained model without fine-tuning, following standard protocol~\citep{caballero2017real}.

\subsection{Experiment Settings}
For Vimeo-90K~\citep{xue2019video}, we use 14-frame sequences by reversing and concatenating the original 7-frame clips. 
For REDS~\citep{nah2019ntire}, we extract 16-frame clips from the original 100-frame sequences. Optical flow is estimated using SpyNet~\citep{ranjan2017optical}, where the network is frozen for the first 5{,}000 iterations 
and then jointly trained with a learning rate scaled to $0.125\times$ of the main model. For feature alignment and resampling, we adopt the implicit alignment module from IART~\citep{xu2024enhancing}. We train our model using the Adam optimizer~\citep{kingma2014adam} with $\beta_1{=}0.9$, $\beta_2{=}0.99$ and the Cosine Annealing Scheduler~\citep{loshchilov2016sgdr}. 
The initial learning rate is set to $2 \times 10^{-4}$, and we use a mini-batch size of 8. Our training is conducted in two stages. 
We first train on REDS for 600{,}000 iterations using 8 NVIDIA H100 GPUs, 
and directly use this model for REDS evaluation. 
The REDS-trained weights are then used to initialize training on Vimeo-90K, 
where we train for an additional 300{,}000 iterations using 8 NVIDIA V100 GPUs, 
and use the resulting model for evaluation on Vimeo-90K-T. Our model configuration is as follows: the embedding dimension is 192 with a propagation depth of four stages. 
Each stage contains two shifted-window self-attention (SWSA) blocks (with and without shift) followed by two GSM blocks. 
SWSA uses 8 attention heads with a window size of $(2, 8, 8)$. 
The window propagation module (WPM) operates on a temporal window of 5 frames (${K}=5$, two past, anchor, two future). 
For the SS2D component of GSM, the state dimension $d_\text{state}$ is set to 16.

\section{Results}\label{results}

\begin{table}
\centering
\small
\renewcommand{\arraystretch}{1.2}
\begin{tabular}{l|c|c|c|c}
\toprule
\multirow{2}{*}{Method} 
  & Param. & FLOPs & Runtime & PSNR \\
  & (M)    & (T)   & (ms)    & (dB) \\
\midrule
VRT & 35.6 & 1.37 & 1394 & 32.19 \\
RVRT & 10.8 & 2.21 & 743* & 32.53 \\
IART & 13.4 & 2.51 & 1703 & 32.90 \\
\textbf{GSMamba} & 10.5 & 1.52 & 1070 & 32.69 \\
\bottomrule
\end{tabular}
\caption{Comparison of model parameters, FLOPs, runtime, and PSNR. Our GSMamba has lower parameter count and FLOPs, achieves shorter inference time, and delivers competitive PSNR compared with other VSR methods. 
(*) Runtime of RVRT is measured with custom CUDA kernels provided by the authors.}
\label{tab:comparison_params_flops_runtime}
\end{table}

\begin{table*}
\centering
\large
\begin{tabular}{c|c|c|c|cc}
\toprule
\textbf{Alignment} & \textbf{Scanning} & \textbf{Anchor} & \textbf{Scatter} & \textbf{PSNR} & \textbf{SSIM} \\
\midrule
\multirow{3}{*}{Align ×} 
  & Spatial-first & – & – & 30.70 & 0.8716 \\
  & 3D Hilbert & – & – & 30.68 & 0.8710 \\
  & Temporal-first & – & – & 30.55 & 0.8666 \\
\midrule
\multirow{3}{*}{Align o} 
  & \multirow{3}{*}{Temporal-first} 
    & Forward & X & 31.74 & 0.8942 \\
  & & Center & X & 31.83 & 0.8950 \\
  & & Center & O & \textbf{31.93} & \textbf{0.8957} \\
\bottomrule
\end{tabular}
\caption{Ablation of scanning and alignment strategies. 
Scanning defines the 1D token ordering for Mamba, 
while Anchor indicates temporal alignment (Forward/Center) within each window. 
The last three rows correspond to the configurations visualized in Fig.~\ref{fig:windowed_propagation} (A–C).}
\label{tab:ablation}
\end{table*}

\begin{figure*}
    \centering
    \includegraphics[width=\linewidth]{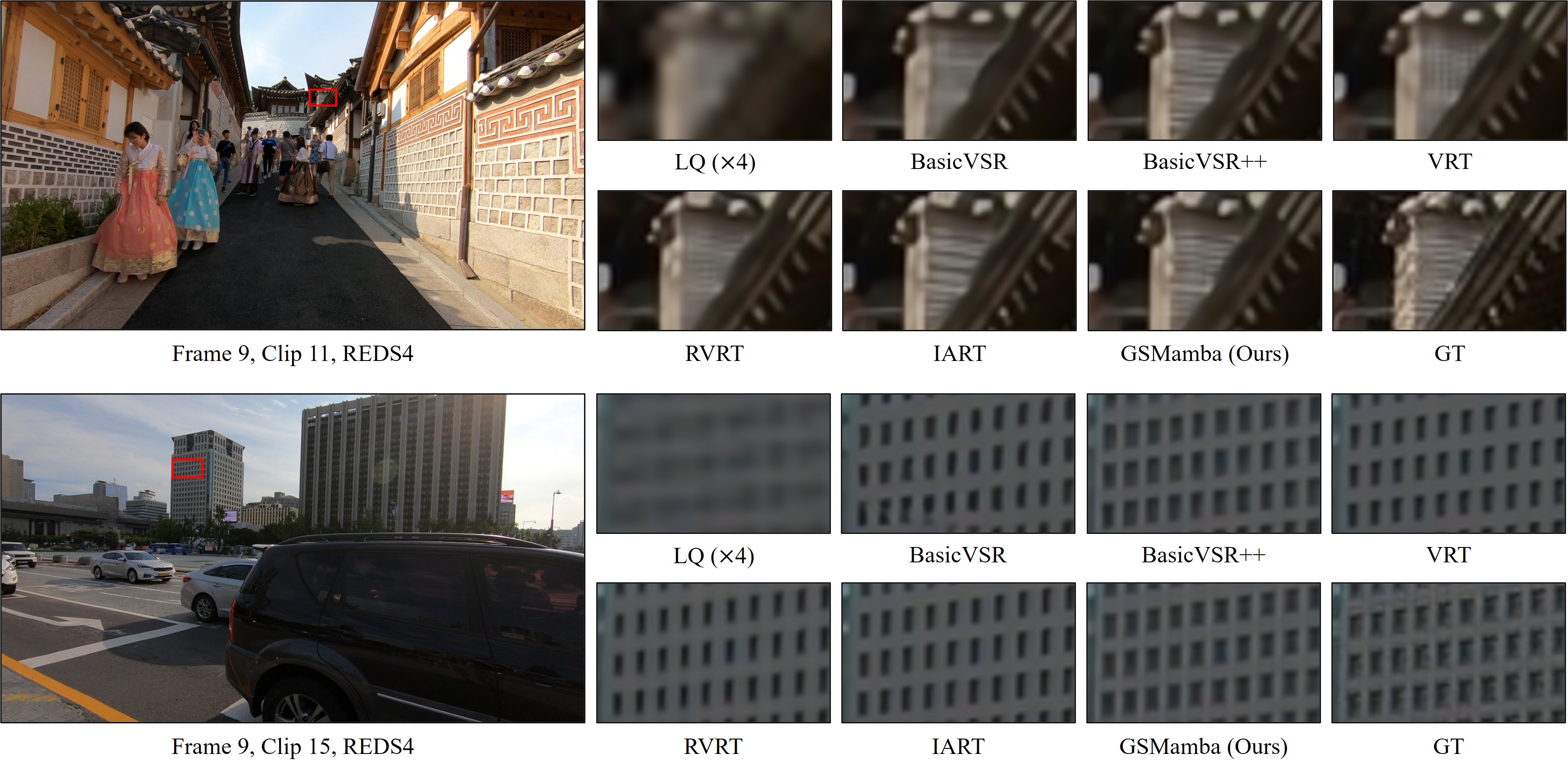}
    \caption{Qualitative results with the state-of-the-art methods on REDS4 \citep{nah2019ntire} dataset}
    \label{fig:architecture}
\end{figure*}

\begin{figure*}
    \centering
    \includegraphics[width=\linewidth]{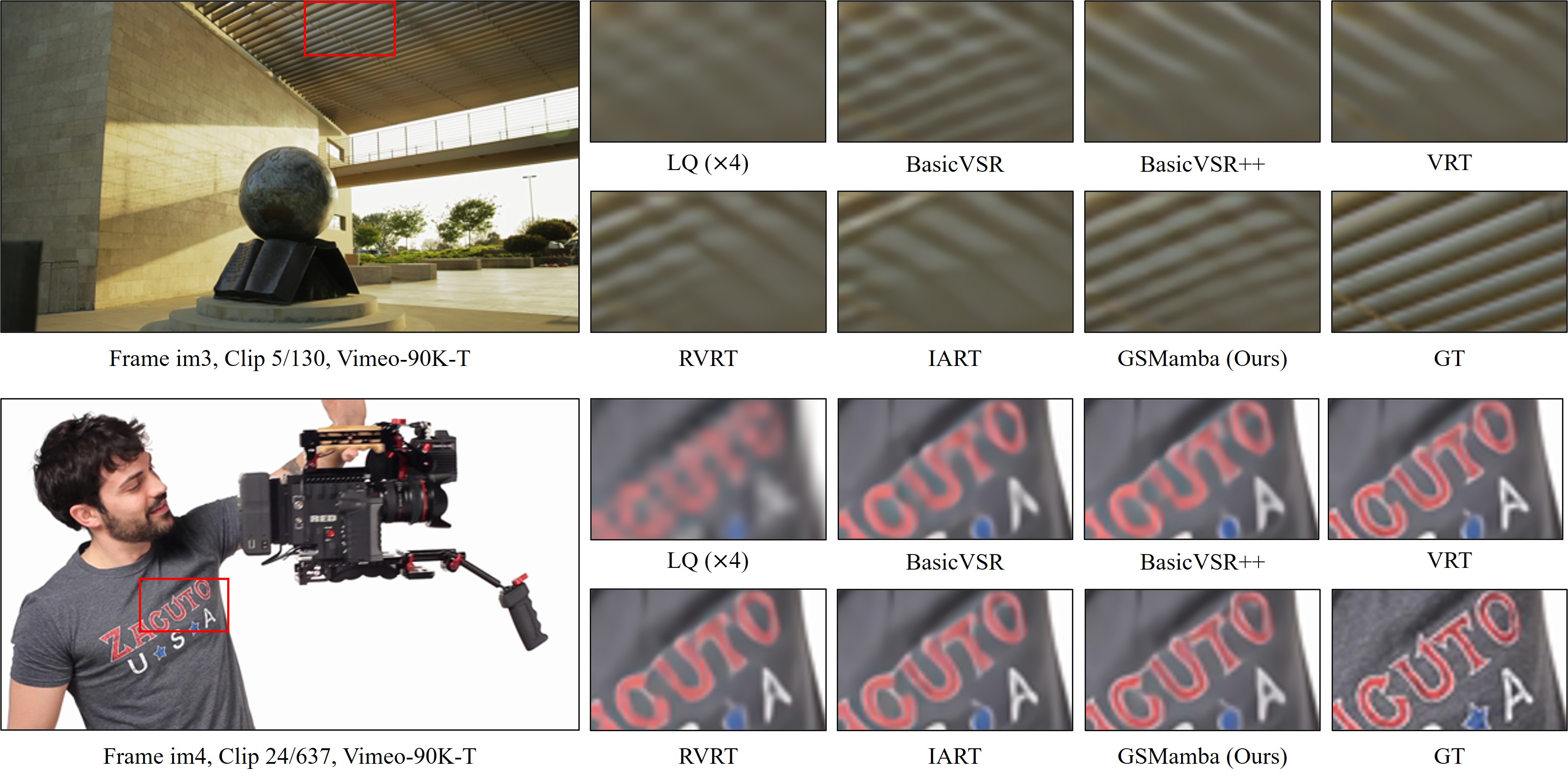}
    \caption{Qualitative results with the state-of-the-art methods on Vimeo-90K-T \citep{xue2019video} dataset}
    \label{fig:architecture}
\end{figure*}

\subsection{Quantitative Results}
We compare PSNR and SSIM metrics on three benchmark datasets against representative VSR baselines, including TOFlow~\citep{xue2019video}, EDVR~\citep{wang2019edvr}, VSR-T~\citep{cao2021video}, BasicVSR, IconVSR~\citep{chan2021basicvsr}, BasicVSR++~\citep{chan2022basicvsr++}, VRT~\citep{liang2024vrt}, RVRT~\citep{liang2022recurrent}, and IART~\citep{xu2024enhancing}. 
The results are summarized in Table~\ref{tab:vsr_comparison}. 
Our method achieves state-of-the-art performance on the Vimeo-90K-T dataset and competitive performance on the REDS4 and Vid4 datasets. 
Combined with the results in Table~\ref{tab:comparison_params_flops_runtime}, 
these findings demonstrate that our approach attains strong reconstruction quality while maintaining high efficiency in terms of parameters and FLOPs.

\subsection{Qualitative Results}
We further provide qualitative comparisons on the REDS4 and Vimeo-90K-T datasets to visually assess the reconstruction quality of our method. 
We compare against strong baselines, including BasicVSR, BasicVSR++, VRT, RVRT, and IART. 
Representative visual results are shown in Figure~\ref{fig:architecture}. Our method produces sharper textures and fewer artifacts, demonstrating superior perceptual quality compared with existing approaches.

\subsection{Model Efficiency}
Table~\ref{tab:comparison_params_flops_runtime} compares the number of parameters, FLOPs, and runtime of our method with other VSR baselines. 
Our model has a lower parameter count, lower FLOPs, and faster runtime compared with existing methods. 
The experiments are conducted on an RTX~3090 GPU using low-quality inputs of $16 \times 320 \times 180$ from the REDS dataset.

\section{Ablation Study}
We conduct ablation studies on the REDS dataset to evaluate the impact of scanning order, the scatter mechanism, and anchor selection in our Gather-Scatter Mamba (GSM). For all ablation training, we extract 6 consecutive frames and use explicit bilinear resampling for alignment during the gather stage (instead of the implicit warping module~\citep{xu2024enhancing} used in the main experiments.) 
The results are summarized in Table~\ref{tab:ablation}.

\paragraph{Scanning Strategy.}
Temporal-first scanning yields the best performance among different token orderings. 
This result highlights that when using Mamba for temporal propagation, explicit alignment becomes necessary and arranging tokens along the temporal axis enables Mamba to more effectively capture long-range dependencies.

\paragraph{Scatter Mechanism.}
Finally, enabling the scatter phase—which redistributes the Mamba outputs back to their original temporal locations—yields additional PSNR and SSIM improvements. This confirms that jointly updating all frames within a window leads to more consistent temporal propagation compared to updating only the anchor frame.

\paragraph{Anchor Selection.}
Center-anchored alignment consistently outperforms forward-anchored alignment. Center anchoring provides two advantages: (i) it reduces occlusions, allowing more nearby frames to contribute to reconstruction, and (ii) it minimizes warping errors due to shorter motion paths, resulting in improved alignment and overall reconstruction quality.

\section{Conclusion}\label{conclusion}

In this work, we propose Gather-Scatter Mamba (GSM), a novel video super-resolution framework that integrates Mamba-based temporal propagation with alignment-aware residual redistribution. Our gather-scatter design ensures that all frames within a temporal window are enhanced jointly, improving both efficiency and consistency. Extensive experiments on REDS, Vimeo-90K, and Vid4 demonstrate that GSMamba achieves competitive or superior performance to state-of-the-art methods while requiring fewer parameters and FLOPs. Moreover, our ablation studies highlight the importance of center-anchored alignment and residual scattering, both of which significantly contribute to the final performance. Our results suggest that structured state-space models such as Mamba are a promising alternative to recurrent or attention-based propagation for video restoration tasks, combining scalability, efficiency, and strong temporal modeling ability.

{
    \small
    \bibliographystyle{ieeenat_fullname}
    \bibliography{main}
}

\end{document}